# Large Language Models for Water Distribution Systems Modeling and Decision-Making

Yinon Goldshtein[1,†], Gal Perelman[2,†,*], Assaf Schuster[3], and Avi Ostfeld[4]

[1] M.Sc. Student, Faculty of Computer Science, Technion – Israel Institute of Technology, Haifa 32000, Israel; E-mail: yinong@cs.technion.ac.il

[2] Ph.D. student, Faculty of Civil and Environmental Engineering, Technion – Israel Institute of Technology, Haifa 32000, Israel; E-mail: gal-p@campus.technion.ac.il

[3] Professor, Faculty of Computer Science, Technion – Israel Institute of Technology, Haifa 32000, Israel; E-mail: assaf@technion.ac.il

[4] Professor, Faculty of Civil and Environmental Engineering, Technion – Israel Institute of Technology, Haifa 32000, Israel; E-mail: ostfeld@technion.ac.il

† Equal contribution

* Corresponding author

## Abstract

The integration of Large Language Models (LLMs) into engineering workflows presents new opportunities for making computational tools more accessible. Especially where such tools remain underutilized due to technical or expertise barriers, such as water distribution system (WDS) management. This study introduces *LLM-EPANET*, an agent-based framework that enables natural language interaction with EPANET, the benchmark WDS simulator. The framework combines retrieval-augmented generation and multi-agent orchestration to automatically translate user queries into executable code, run simulations, and return structured results. A curated set of 69 benchmark queries is introduced to evaluate performance across state-of-the-art LLMs. Results show that LLMs can effectively support a wide range of modeling tasks, achieving 56–81% accuracy overall, and over 90% for simpler queries. These findings highlight the potential of LLM-based modeling to democratize data-driven decision-making in the water sector through transparent, interactive AI interfaces. The framework code and benchmark queries are shared as an open resource.

## Highlights

- We present a no-code framework for natural-language interaction with EPANET
- The framework uses a multi-agent pipeline, RAG, code execution, and self-debugging
- The framework achieves up to 90% accuracy across networks, LLMs, and query types.
- The framework democratizes hydraulic and quality simulations for broader adoption
- We share open-source code and a benchmark set of natural language queries

## Software and Data Availability

- Name of software: LLM-EPANET
- Developers: Yinon Goldshtein, Gal Perelman
- Contact: yinong@cs.technion.ac.il
- Date first available: September 2025
- Software required: Python 3.10, OpenRouter account, Docker
- Program language: Python
- Source code at: https://github.com/yinon-gold/LLMs-in-WDS-Modeling
- Documentation: Provided in the README.md file within the OSF archive

**Keywords:** LLM, AI Democratization, Human Computer Interactions, Retrieval Augmented Generation, Code Generation, EPANET

## 1. Introduction

Water Distribution Systems (WDS) are critical infrastructure systems essential for the sustainability of human societies. With accelerating urbanization, aging infrastructure, and growing climate variability, the effective management of these systems is more important than ever. To address this challenge, varied computational tools and simulation models have been developed to optimize system performance, predict hydraulic and water quality dynamics, and support decision-making processes (Fu et al., 2024). However, despite their potential, the adoption of these computational tools is often constrained by the significant technical expertise required to operate them. Key challenges that were recognized as a burden to the broader adoption of computational tools are the limited trust in models' reliability, a lack of transparency, and the considerable time and effort invested in model setup, calibration, and maintenance (Samadi et al., 2024; Vekaria and Sinha, 2024). Furthermore, the expert knowledge that is required to interact with models means that model-based decision-making is limited to stakeholders with high levels of technical skills (Rao and Salomons, 2007).

A prominent example of such a computational tool is the Environmental Protection Agency Network Evaluation Tool (EPANET), developed by the United States Environmental Protection Agency (EPA). EPANET is a widely used open-source software tool that simulates the hydraulic and water quality behavior in WDS (Rossman, 2000). With more than 4,500 citations in ranked journals and hundreds of thousands of downloads, EPANET has become a benchmark in WDS modeling. While originally written in C programming language, the software has been extended through several open-source wrappers in varied programming languages (Arandia and Eck, 2018; Eliades et al., n.d.; Klise et al., 2017; Kyriakou et al., 2023), making it highly accessible for engineers and scholars. Its robust simulation capabilities have made EPANET a cornerstone in water engineering, supporting critical applications such as leak detection (Levinas et al., 2021), optimal energy management (Luna et al., 2019), sensor placement optimization (Cardoso et al., 2021; Ohar et al., 2015), water quality control (Elsherif et al., 2023; Perelman and Ostfeld, 2025), and cyber attack mitigation (Taormina et al., 2019). However, the practical use of EPANET remains largely confined to technically skilled users. The typical end user relies on the software GUI to interact with simulations, whereas more advanced usage usually requires technical coding skills, hindering broader adoption by non-technical water sector employees (Sela et al., 2019). Even the GUI-based usage is often limited within water utilities, where engineers and decision-makers are occupied with urgent operational tasks, leaving little time or incentive for engaging in model-based analysis. Consequently, the valuable insights offered by sophisticated computational models such as EPANET are often underutilized, restricting decision-making to a narrow group of experts.

Recent advancements in Artificial Intelligence (AI), and particularly in Large Language Models (LLMs), have created new opportunities to engage water sector professionals with data-driven decision-making. LLMs have demonstrated considerable promise in code generation tasks, making them promising intermediaries between humans and models. By translating plain language prompts into executable commands, LLMs enable intuitive interactions with sophisticated computational tools, thereby lowering technical barriers and expanding access to critical technical insights.

Initial efforts to explore the integration of LLMs into water resources management have been presented in several recent studies. Sela et al. (2024) surveyed various workflows in water utilities and demonstrated how generative AI can support tasks ranging from missing data imputation to asset management and demand analysis. Their findings highlight the value of natural language interfaces in reducing the technical burden traditionally associated with data analysis. Marzouny and Dziedzic (2024) proposed a few-shot learning strategy for pump scheduling where the LLM iteratively generates pumping strategies, receives feedback from EPANET simulations, and refines its suggestions in subsequent iterations. Taormina and Werf (2024) developed an LLM-based framework for pipe defect detection using image data. In addition to identifying faults, their model provides interpretability through language-based explanations of its decisions.

The potential of LLMs for WDS can also be learned from cross-domain use cases. For instance, LLM is utilized for flood risk management by allowing stakeholders to query databases and visualize scenarios (Martelo et al., 2024), and by reasoning about flood causes (Zafarmomen and Samadi, 2025). In the power grid domain, LLMs have been used for power flow simulations (Jia et al., 2024) and supporting long-term planning decisions (Chaturvedi et al., 2025).

Despite the promising potential showcased in the above studies, a fully integrated framework that enables water professionals to freely interact with hydraulic and water quality simulations across a broad range of analytical and operational tasks remains absent. Developing such a domain-specific, no-code framework requires addressing three key challenges: (1) generating accurate domain-specific code tailored to complex simulation environments like EPANET; (2) creating a seamless, no-code, prompt-to-execution pipeline that eliminates manual intervention; and (3) designing transparent, trust-building interfaces that empower non-technical stakeholders to understand and verify how their inputs influence simulation outcomes. The following subsections elaborate on each of these challenges and introduce our proposed LLM-based framework for EPANET, which is expressly designed to bridge these limitations and democratize access to advanced water system modeling.

### 1.1. Customized Code Generation

LLMs have revolutionized code generation by leveraging vast amounts of publicly available source code and technical documentation. Tools such as OpenAI's Codex (Chen et al., 2021) and GitHub Copilot (GitHub Copilot, 2023) have shown impressive capabilities in generating high-quality code using widely adopted programming libraries, benefiting from the abundance of training data available on the internet (Austin et al., 2021; Chen et al., 2021). However, these models are primarily trained on mainstream programming examples, leaving a notable gap when it comes to niche domains. In specialized fields, like water distribution system simulations, the scarcity of domain-specific code in the training data limits the models' ability to generate effective, tailored solutions (Joel et al., 2024). This gap is particularly pronounced in environments like EPANET, where simulations rely on complex, structured inputs and nuanced domain knowledge.

### 1.2. Integration

Most existing LLM-based code generation tools assume a human-in-the-loop workflow, where users manually review, modify, and execute the suggested code. While suitable for technical users, this interaction model is impractical for non-experts. A no-code framework requires a fully automated workflow in which natural language prompts are converted into validated, executable code and run seamlessly within the simulation environment. This presents new technical challenges related to syntax validation, error handling, and runtime evaluation (Wang et al., 2024). Recent work in agent-based systems offers promising directions: Zhuge et al. (2024) proposed an "Agent-as-a-Judge" architecture, where autonomous agents evaluate the outputs of peer agents. Similarly, Li et al. (2023) introduced a multi-agent collaboration framework for code generation, validation, and iterative improvement in optimization contexts. The integration part is recognized as a key aspect for extending the user base. Our proposed approach directly addresses this need by automating the prompt-to-execution workflow, thereby laying the groundwork to engage users with AI-based tools.

### 1.3. Reliability, Usability, and Trust

Ensuring both usability and trust is pivotal for making complex simulation tools accessible to non-expert users (Samadi et al., 2024; Vekaria and Sinha, 2024). Beyond generating functionally reliable code, it is essential to also communicate how natural language inputs are transformed into simulation commands. Such transparency, potentially through the display of the generated code with accompanying explanations or documentation comments, enables users to verify and understand the underlying processes, thereby fostering confidence in the system. Equally important is the interpretability of simulation outputs; when users can readily grasp the underlying computations and assumptions, they are more likely to trust the tool's recommendations. By addressing these factors, the proposed framework minimizes potential errors and ambiguities in code generation and cultivates a robust, user-centric experience that promotes broader adoption within the water sector.

The proposed framework aims to close these gaps by developing an LLM-based interface for EPANET. The primary objective of this framework is to democratize access to advanced simulation tools and engage non-technical stakeholders toward a culture of data-driven decision-making based on advanced software adoption at all levels of water utilities.

## 2. Methodology

As explained above, LLMs excel in generating human-like texts, including programming code. This advancement has significantly impacted software development, automating coding tasks and aiding developers in writing complex code more efficiently. The code generation capabilities of LLMs stem from their training on diverse datasets compiled from publicly available code repositories, enabling them to learn coding patterns and language semantics from vast amounts of data (Bommasani et al., 2021). However, in more specialized applications, the accuracy and context-specific relevance of the code are critical. LLMs can sometimes generate code that is syntactically correct but functionally inaccurate or irrelevant, a challenge that is amplified when dealing with specialized niche software. To mitigate these issues, approaches like Retrieval-Augmented Generation (RAG) have been proposed, which enhance the code generation process by integrating targeted retrieval of relevant code snippets from a specific codebase (Lewis et al., 2020). The essence of RAG is to combine the generative capabilities of LLMs with an external search system that retrieves relevant information (like specific software documentation) to focus the LLM's attention on the most relevant parts that will help generate an accurate answer. This method ensures that the generated code is not only correct but also contextually appropriate for the task at hand.

The proposed methodology utilizes an RAG technique to develop the LLM-EPANET framework. It is designed to convert natural language queries into executable Python code, customized to run any required simulation, providing a sufficient answer to the queries. The interactions with the EPANET engine are done through the Python wrapper EPyT (Kyriakou et al., 2023), where the package documentation serves as an external knowledge base that is combined with the generative LLM capabilities. We use RAG to retrieve the EPyT documentation and convert it into a numerical multi-dimensional structure that allows computations. The goal is to create a numeric database that captures the relationships between different parts of the original textual data, a structure referred to as a vector database (VDB). For example, based on the vectorized data, we can identify and quantify similarities between the user query and parts in the EPyT documentation that can help answer this query. This process is usually referred to as vector embeddings and is generally presented in Figure 1.

## 2.1. RAG Preprocess

The RAG implementation begins with preprocessing the EPyT Python package documentation. As our evaluation did not include the Multi-Species Extension (MSX) features of EPANET, all MSX-related functions were excluded. The EPyT documentation often includes a section at the end of each function description listing other related functions, typically introduced with the phrase “See also”. These sections are intended to help human users navigate between related parts of the codebase. However, in our vector database, each function description is treated as an independent entry. Including these cross-references can introduce unrelated function names into the text, which may confuse the similarity search process and lead to irrelevant results. Therefore, we removed all "See also" sections during preprocessing to ensure cleaner and more accurate retrieval.

Additionally, to improve retrieval quality and ensure completeness, missing function docstrings were automatically generated using a commercial LLM. Each function, including its signature, arguments, and accompanying description (original or generated), was treated as a single semantic unit. These function level text entries were used as the basic chunks for indexing in the vector database. Unlike token-based chunking schemes, our approach uses function-level granularity, resulting in non-overlapping entries and preserving semantic cohesion across the indexed knowledge base.

### 2.2. Embeddings Model

For semantic embedding, we evaluated several state-of-the-art code-oriented embedding models, including Voyage Code 3, OpenAI text-embedding-3-large, and NVIDIA NV-Embed-v2. These models produce dense vector representations with dimensionalities ranging from 2048 to 4096, using 32-bit floating-point precision. Embeddings were stored at full dimensionality without dimensionality reduction. Semantic similarity between queries and indexed documentation entries was computed using cosine similarity. At inference time, the top-ranked function-level entries were retrieved from the vector database and provided to the LLM as contextual grounding for code generation. Comparable retrieval behavior was observed across all the tested models. Therefore, embeddings were treated as a fixed infrastructure component rather than an object of optimization in this study, isolating the effect of model choice and prompt design from retrieval configuration. For the final implementation, we selected VoyageAI's *voyage-code-3* model, which at the time achieved state-of-the-art performance on the Massive Text Embedding Benchmark (MTEB) (Muennighoff et al., 2022) for retrieval tasks. Other tested models included OpenAI's *text-embedding-3-large* and NVIDIA's *nv-embedqa-mistral-7b-v2*. The embedding model was used to transform each function chunk into a fixed-dimensional dense vector representation.

### 2.3. Similarity Search

The resulting embeddings were stored in a vector database using Facebook AI's FAISS

library (Douze et al., 2024), accessed through the LangChain framework. We used Euclidean (L2) distance as the similarity metric, although experimentation with other metrics, such as L1 distance, revealed minimal differences in retrieval outcomes. During query time, the user's prompt is embedded using the same embedding model and compared to all stored function vectors.

The retrieved documentation chunks are combined with the user query and a task-specific meta prompt to form the full input for the LLM. A description of this full sequence is below. This setup enables high-quality, domain-grounded code generation without requiring the LLM to memorize or guess EPyT syntax, thereby improving interpretability and reducing hallucination risk.

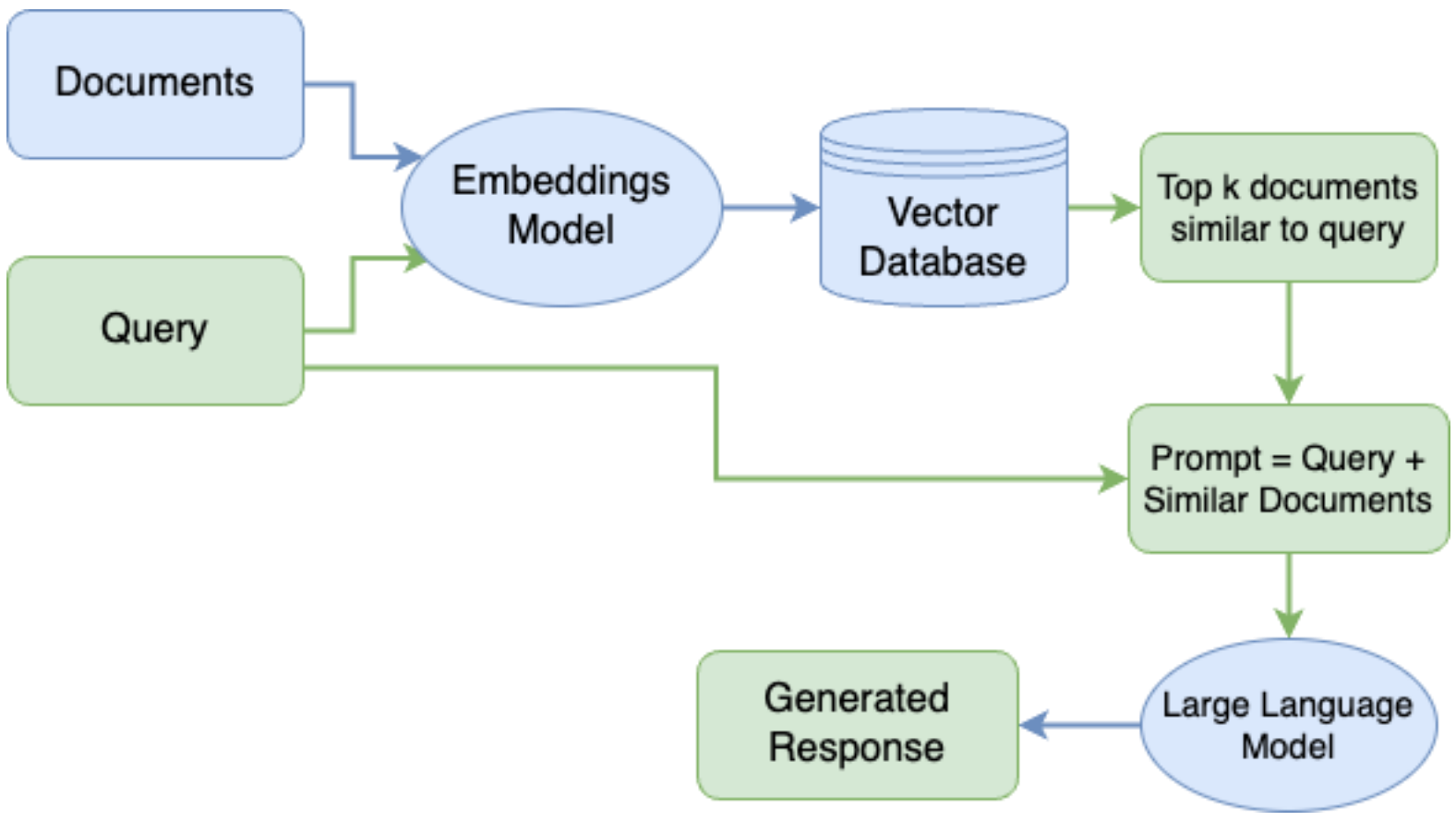

**Figure 1:** *A Simple RAG Workflow*

## 2.4. <u>Semantic Retrieval Validation</u>

To ensure reproducibility and to validate the quality of the retrieval component, we quantitatively evaluated the proposed semantic search mechanism. Retrieval performance was assessed using precision, recall, and F1 scores as a function of the number of retrieved functions ($k$), averaged across all queries. Ground truth relevance for retrieval evaluation was defined retrospectively based on successful executions. Specifically, after completing all experiments, we extracted the EPANET functions that were actually invoked by the generated code for queries that produced correct answers. For each query, the union of functions used across all successful runs was treated as the set of relevant functions. Retrieval performance was then evaluated at the function level by comparing the Top-K functions returned by the RAG component ($k$ = 5) with this ground-truth set, and computing precision, recall, and F1 scores for each query.

Figure 2 (a) shows the expected trade-off between precision and recall as *k* increases, resulting in a shallow optimum in mean F1 at intermediate values of *k*. To assess robustness across queries, Figure 2 (b) reports the distribution of per-query F1 scores for each *k*, highlighting increased variability for very small k and the best median in k=5. Based on this analysis, we fixed *k=5* for all experiments. An additional ablation study comparing semantic retrieval with a hybrid lexical–semantic search strategy is reported in the Supplementary Materials (Figures S1 and S2). A sensitivity analysis for k is in Supplementary Material S3.

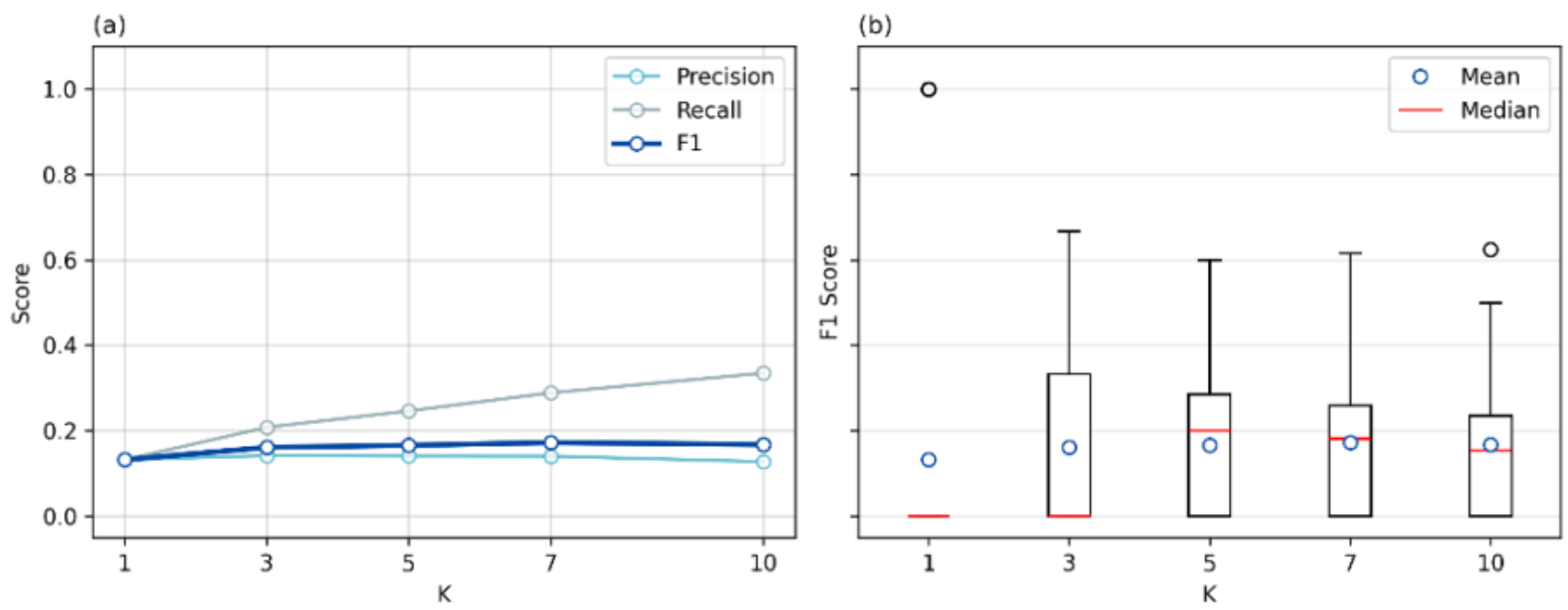

**Figure 2:** *Quantitative evaluation of semantic retrieval performance. (a) Mean precision, recall, and F1 score. (b) Distribution of per-query F1 scores for each k.*

### 2.5. Multi-Agent LLM-EPANET Framework

Next, the main pipeline, LLM-EPANET, is developed as a multi-agent system organized to complete the different tasks required for providing the user with an answer within a no-code process. The following describes the process and assignments of the different agents involved in each stage. The numbering points to the relevant parts on the framework graphic description presented in Figure 3.

The process is triggered by a natural language query from a user (1). The query is transformed into a VDB with the same embedding model used for the EPyT documentation (2). Once the user's query is parameterized with the same VDB as the EPyT codebase (3), mathematical similarity search can be used to focus the LLM on the most relevant parts that are likely needed to generate a code that will answer the query. The similarity search is done based on the Top-K retrieval technique (Lewis et al., 2020), where the most relevant *K* functions of EPyT are retrieved from the documentation, where in this study, *K=5* was used based on several trials and errors. Management and similarity search through the VDB are done with Facebook AI

Similarity Search (FAISS) (Douze et al., 2024). To construct the final prompt that will be sent to the LLM, the user query and the top $K$ functions context are joined with a meta prompt that provides the LLM with specific context (4). The meta prompt includes a general description of the task and specific instructions important for working with EPyT package and EPANET. The design of this meta prompt was found to be crucial for achieving good performance; see further details below.

The crafted prompt is sent to an LLM, which generates a code block of a Python function definition constructed from the EPyT functions to build a calculation that will answer the requested information (5). The Python function generated by the LLM and the requested task query are inserted into another pre-constructed prompt and sent to a smaller language model to construct a one-line command that evaluates the generated function (6) and is combined into a Python script that first initiates an EPANET network object and then executes the code generated by the LLM. This script is then compiled and run in an isolated environment and returns the result variable (7). When the script executes successfully without errors, the output of the execution block is returned to the user (8). In the case of a failure due to a code error, the generated function definition and evaluation script, along with the error traceback, are sent back to the LLM as a structured error handling prompt (9) and then returned to step (5) to rewrite a new function code block to address the error. This process is repeated iteratively up to a pre-defined number of debugging retries, configured by the user, to ensure as robust and error-free an implementation as possible.

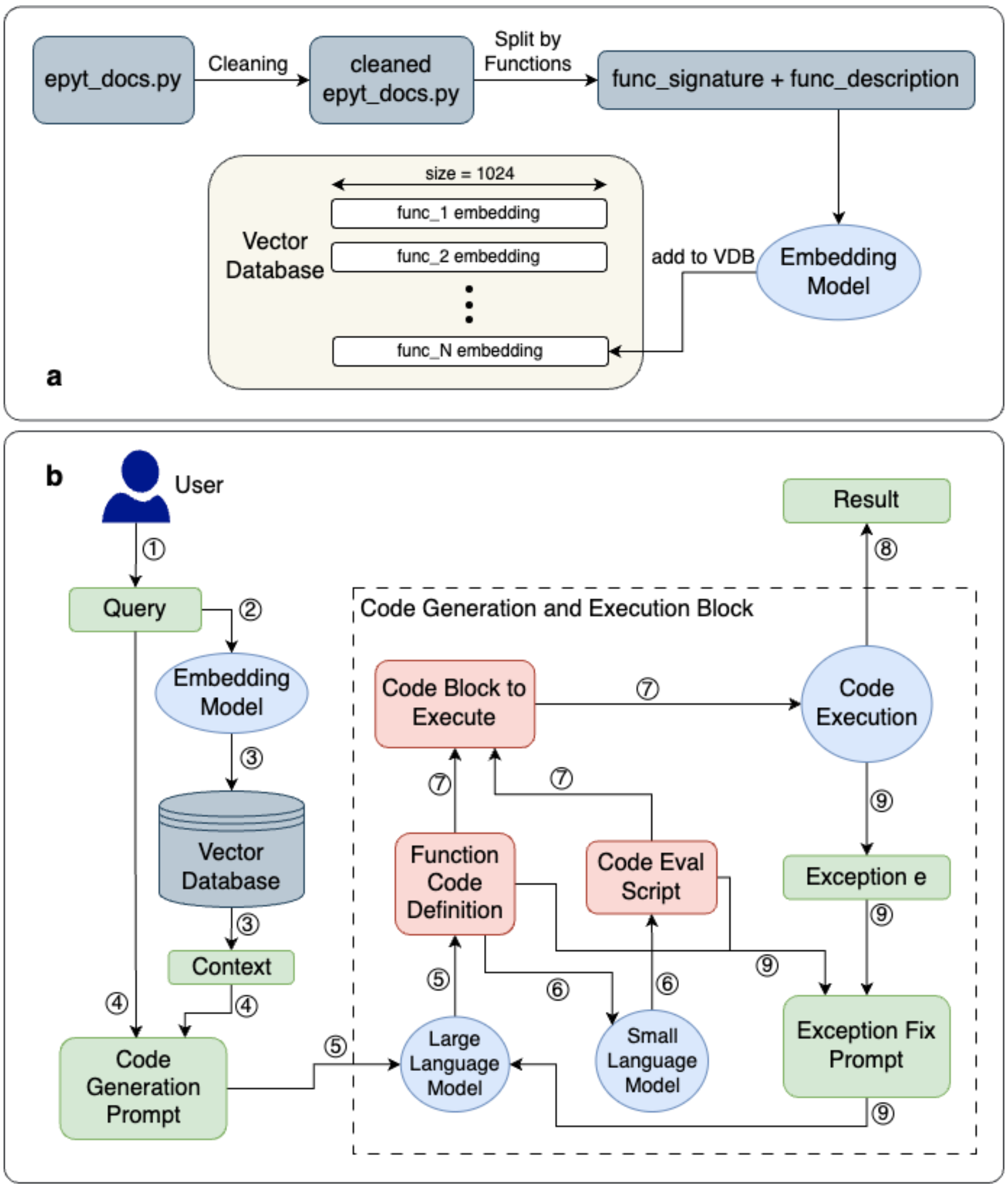

**Figure 3**: *Architecture of LLM-EPANET. **(a)** Preprocessing steps for creating the vector database from the EPyT documentation. **(b)** Illustration of LLM-EPANET pipeline workflow.*

For the communication between agents, the LLM-EPANET framework employs a sequential, orchestrated multi-agent pipeline, not involving parallelism or autonomous agent negotiation. Each agent operates deterministically on the output of the previous stage, with intermediate results passed explicitly through structured data objects while not sharing mutable memory. Retrieved documentation chunks, generated code, execution outputs, and error traces are serialized and injected into subsequent prompts as contextual input. At each stage, a single agent is responsible for producing the next action (e.g., code generation, command construction, or error correction). Conflicts between alternative agent outputs do not arise, as only one candidate solution is generated and evaluated per iteration.

The last steps of the above sequence present the mechanism of handling code errors. When generated code fails to execute, the Python interpreter returns a structured traceback containing the error type and message. This traceback, together with the previously generated code and the original task description, is provided as input to the code-generation agent in the next iteration. The agent is instructed to revise the code specifically to address the reported error. The self-debugging process is governed by a predefined maximum number of retries, configured by the user (set to 5 retries in this study, see Robustness Analysis section below). Execution terminates successfully when the code runs without errors and returns an output, or unsuccessfully when the retry limit is reached. No explicit convergence criterion beyond successful execution is imposed, reflecting the goal of obtaining runnable, task-complete code rather than optimizing intermediate representations. In terms of AI safety, our framework is designed to enable LLM-generated code to be executed in an isolated environment. The execution can be configured to run inside a containerized sandbox (Docker-based) via an optional "*sandbox*" flag. When this flag is enabled, code is executed within a sandbox session with an explicit timeout (60 seconds in our experiments) and a restricted set of dependencies limited to the libraries required for simulation (e.g., NumPy and EPyT). The EPANET input files are copied into the isolated environment before execution, and not accessed from the host system. Execution outputs are captured and returned to the framework, and failures (timeouts or uncaught exceptions) trigger the self-debugging loop subject to the predefined retry limit.

### 2.6. Meta Prompt

The meta prompt is a general-purpose instruction appended to every query to improve the context provided to the LLM (see step 3 in Figure 3). For example, the general description starts with: "*You are a code assistant for a water engineer. Your task is to write code snippets that interact with the EPyT Python package based on its documentation*". Following this declaration, instructions on how to run EPyT can be provided with different levels of detail, according to the specific use case.

As discussed earlier, one of the key challenges that motivated this study is the generation of domain-specific customized code, particularly given the limited representation of specialized packages in LLM training datasets. The meta prompt is designed to mitigate this limitation by providing the LLM with targeted modeling guidance and package-specific syntax patterns. Conceptually, the meta prompt can be viewed as an expert modeler offering real-time advice to a junior engineer. It contains concise but critical modeling tips tailored to EPANET and EPyT, ensuring that the LLM can produce accurate and context-aware code. For example, EPANET assigns both an integer index and a string ID to each network element, a distinction that can be confusing, especially for beginners using coding interfaces that abstract these differences. To prevent related errors, the meta prompt explicitly instructs the LLM on the correct usage of these identifiers, as illustrated in Appendix A.

The meta prompt was developed iteratively through empirical testing across diverse query types, with modifications introduced only when recurrent and systematic failure modes were observed. The impact of additional instructions was validated through a comparison against previous versions of the meta prompt and the base prompt.

While the use of a detailed meta prompt might raise concerns about over-reliance on injected domain knowledge, we emphasize that this is an intentional design choice rather than an attempt to demonstrate autonomous reasoning. The goal of the framework is not to replace experts' work, but to embed best-practice modeling knowledge into a natural-language interface, enabling non-expert users to interact safely and productively with established simulation tools. The meta prompt provides general guidance, not aimed at specific tasks, analogous to the documentation or onboarding a human modeler would receive when first interacting with EPANET through code. The LLM is still required to parse, interpret, and translate diverse natural language queries into structured code. In this sense, the meta prompt serves as an instructor that enhances the reliability of the code and reduces hallucinations. Moreover, the meta prompt concept opens up opportunities for sophisticated prompt management in applied and commercial settings. Organizations developing LLM-augmented modeling tools can maintain collections of task-specific prompts, optimized for specific use cases. These prompts can evolve over time, through version control, performance tracking, and user feedback. Thus, forming a modular infrastructure that tailors LLM behavior to operational needs without requiring model retraining. To further examine the importance of the meta prompt, we also tested a reduced baseline version that includes only minimal task instructions and excludes domain-specific modeling tips as detailed in the Results section below.

## 2.7. Evaluation

To evaluate the performance of our framework, we curated a list of natural language queries designed to test its capabilities across various aspects of WDS applications. These queries covered a range of desired interactions with the EPANET software. The objective of the framework is to produce correct simulation outputs rather than text or code matching. Hence, the evaluation is based on the functional correctness of the structured results rather than graded textual similarity metrics. For each benchmark query, a reference implementation was manually written using EPANET through the EPyT Python interface and executed before evaluating LLM-generated outputs. These reference implementations are provided as executable test scripts. Each script runs the appropriate simulation and produces a deterministic reference output, which serves as the primary ground truth for evaluation. This approach ensures that reference results are reproducible, transparent, and independent of the LLM-generated code.

Correctness evaluation was done by comparing the LLM-generated outputs against the reference outputs produced by the test scripts. A response was labeled *correct* if the returned value was equivalent to the reference result. Code execution failures, unrecovered errors, incorrect aggregation, indexing mistakes, or unit mismatches were labeled as *incorrect*. In a small number of cases, LLM-generated outputs slightly differed from the reference results while still being semantically correct. Such cases arose due to minor numerical discrepancies (e.g., rounding or differences in simulation execution pathways within EPyT) or alternative but equivalent representations of the same answer (e.g., scalar versus structured outputs). To account for these situations, a hydraulic modeling expert reviewed all mismatches between LLM outputs and reference results and determined correctness based on engineering judgment. The human reviewer did not modify generated code or outputs, but only assessed equivalence relative to the reference task definition.

The queries were categorized by increasing levels of complexity, based on the expected difficulty in understanding the user request and writing appropriate code to provide an answer. The different categories, with example queries, are presented in Table 1.

**Table 1** - Queries Categories and Examples

| Category | Description | Example |
|---|---|---|
| Static | Questions about the network that can be answered without running a simulation | How many pumps are in the network? |
| Hydraulics | Queries regarding the hydraulic dynamics of the network that require an EPANET hydraulic simulation | What is the maximal pressure in the network? |
| Quality | queries related to water quality dynamics, requiring a water quality simulation | What is the average water age in the network based on the last 24 hours of the simulation? |
| Hydraulics Scenario | queries that include network modifications before solving a hydraulic simulation | What will be the max pressure if pipe ID 173 is closed? |
| Iterative | Queries require iteratively changing a parameter in the network and evaluating the impact on another parameter | The pipe with ID 212 needs to be replaced. Available diameters are: [6, 8, 10, 12, 14]. Try to replace the candidate diameters for pipe 122 to answer: What is the minimal diameter that can be used for pipe 122 such that the pressure in all demand junctions will be above 30. Demand junctions are junctions with a base demand larger than 0. To get the base demands of all nodes, use: *d.getNodeBaseDemands()[1]*. To replace a pipe diameter, you can use the EPyT method *d.setLinkDiameter()*, |

Queries were assigned to categories based on the most complex modeling operation required to answer them, as summarized in Table 1. When a query involved multiple steps (e.g., modifying the network and subsequently running a simulation), it was assigned to the category corresponding to the highest-level operation. For example, queries that required network modifications followed by hydraulic simulation were labeled as Hydraulic Scenario, rather than Hydraulics. Category definitions and labels were established prior to evaluation and applied consistently across all models.

It is noted that the iterative category is serving only as an exploratory demonstration of the framework's ability to generate and execute multi-step engineering logic, such as parameter sweeps and conditional feasibility checks. Such tasks go beyond a single query-response paradigm and illustrate how the proposed framework can be integrated into engineering workflows. As such, this category includes only a minimal number of three exploratory queries to illustrate the capabilities of LLM-EPANET beyond a question-and-answer framework. Given the complexity of the iterative queries, their prompts are longer and include more engineering hints about how to code the requested task.

The set of queries refers to three different benchmark networks: Net1 and Net3 from the original EPANET software (Rossman, 2000), and the L-Town network presented by (Vrachimis et al., 2022). In Net3, the water quality analysis is set to trace the Lake source (as in the original network), representing the portion of the Lake water in the network nodes during the simulation. In the L-Town network, the water quality analysis was modified to calculate the water age. The benchmark queries and answers, and ground truth scripts are all shared as open source to facilitate reproducibility and enable future work on natural-language interfaces for EPANET, see Data Availability statement below.

## 3. Results

Based on the LLM-EPANET framework and the set of evaluation queries, several experiments were conducted to examine different configurations of commercial LLMs and parameters. Each query is evaluated with a range of commercial LLMs. However, this study does not aim to benchmark LLM providers. Instead, models were selected to represent different cost-performance regimes to evaluate the robustness of the proposed LLM-EPANET framework. Accordingly, the presented models should not be seen as a comparison between providers, where different levels of models were selected from each provider. Furthermore, given the rapid evolution of LLM capabilities, performance rankings across models, like the one presented in this study, should be interpreted cautiously. The improvements in underlying models are expected to translate directly into improved performance of the LLM-EPANET framework without any architectural changes.

In each experiment, the answers obtained by the LLM-EPANET were compared with expected results in a human-involved review process to determine whether the LLM-EPANET response was correct or not. A query is considered successful if the system's output matches an answer obtained by a simulation that was configured and validated by a human, where cases of no output, such as those caused by unrecovered errors, are considered failures. The accuracy for each category is computed as the percentage of correct answers out of the total queries in the same category. Figure 4 presents an example query from the hydraulic scenarios category for the L-twon network, and the LLM-EPANET output of the generated code and the final numeric answer.

What will be the average pressure in the network if all consumers consumes 10% above the base demand?

```
def calculate_average_pressure_with_increased_demand(d):
    # Get original base demands for all nodes
    original_demands = d.getNodeBaseDemands()[1]
    # Increase demands by 10%
    increased_demands = [demand * 1.1 for demand in original_demands]
    # Set the increased demands
    d.setNodeBaseDemands(increased_demands)
    # Run hydraulic analysis
    d.getComputedHydraulicTimeSeries()
    # Get node pressures
    pressures = d.getNodePressure()
    # Calculate average pressure
    average_pressure = sum(pressures) / len(pressures)
    # Restore original demands
    d.setNodeBaseDemands(original_demands)
    return average_pressure
```

```
46.0348625758192
```

**Figure 4**: *Example for a query from the hydraulic scenarios category, the output code snippet, and the final answer*

A base run tested different state-of-the-art LLMs against the entire set of queries as presented in Figure 5. The results illustrate that the proposed framework can correctly answer a wide range of natural language queries, with accuracy varying across both query categories and models. As explained in the Evaluation section above, complexity refers to the difficulty of translating a query into correct, executable simulation code and aggregating its results. Overall, the accuracy rates are correlated with the categories' complexity and decrease with the increase in complexity. In the static category, where only minimal code is required, such that not involve simulation runs, all the models achieved above 90% accuracy. With the increase in query complexity, the accuracy rates are gradually deteriorating. Nevertheless, the best model (*openai/o3-mini-high*) scored above 60% across all categories. Figure 5 highlights the superiority of certain LLMs over others, aligning with the recent literature on LLM capabilities for code generation and execution (Jiang et al., 2024; Zhuo et al., 2025).

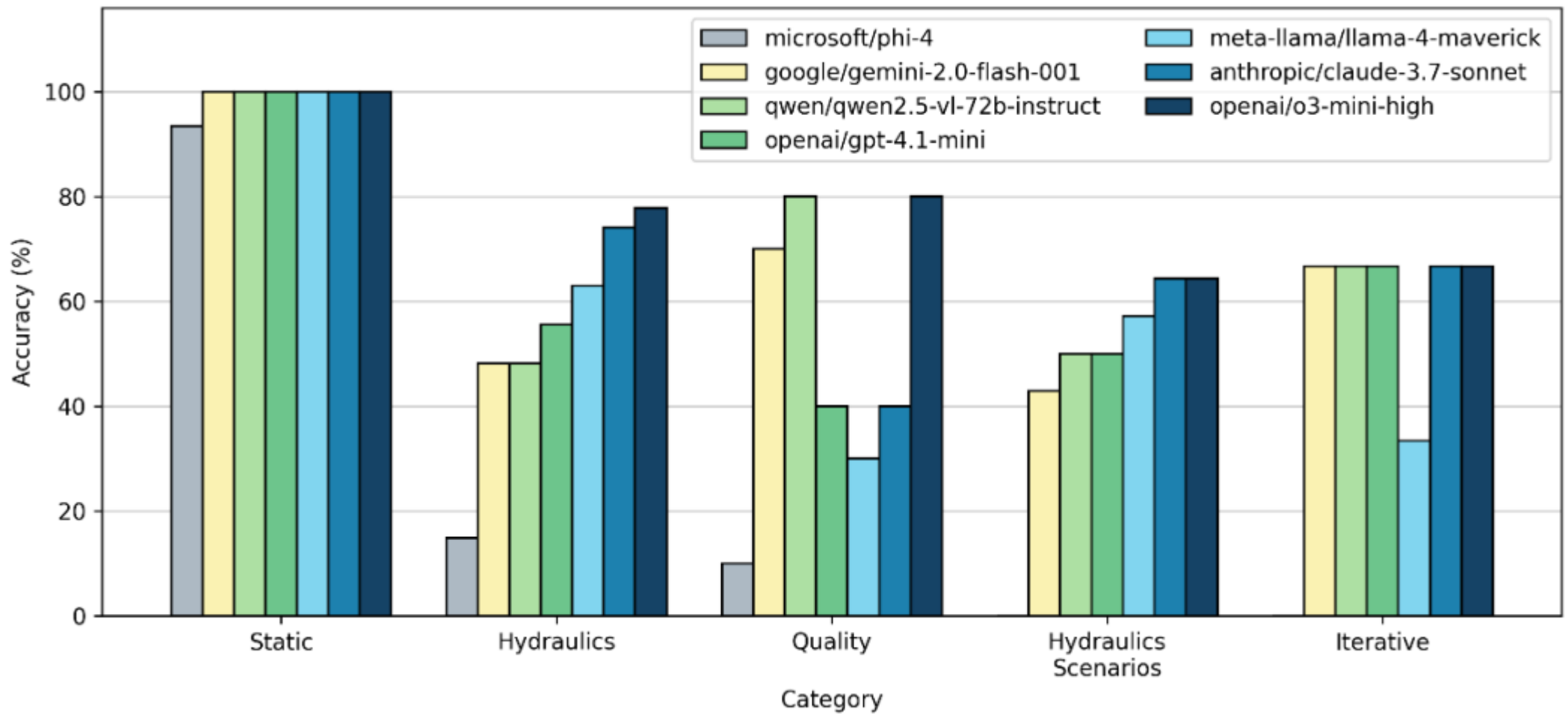

**Figure 5**: *Categorized accuracy of LLM-EPANET results from different LLMs*

The Iterative category differs fundamentally from the other query types evaluated in this study. Instead of mapping a single natural language request to a single executable simulation and aggregated result, these queries require the generation of multi-step control flow, repeated simulations, and conditional logic. They resemble exploratory engineering workflows, such as feasibility checks, parameter sweeps, or design alternatives evaluation. The purpose of this category is not to provide a statistically robust performance estimate, but to demonstrate the feasibility of extending the LLM-EPANET framework beyond single-shot queries toward more complex, process-oriented interactions. The results indicate that modern LLMs can, in some cases, support such workflows successfully, highlighting a promising direction for future research and more systematic benchmarking.

## 3.1. Robustness Analysis

We further assessed the robustness of the framework by analyzing how the overall accuracy evolves as additional debugging attempts are allowed (step 9 in Figure 3). Each additional attempt represents both increased user waiting time and additional cost, as LLMs are billed per token. Figure 6 presents the cumulative accuracy as a function of the maximum number of debugging attempts permitted, reported separately for each model. This presentation indicates the portion of answers that were answered correctly up to the $i^{th}$ attempt, out of the total set of queries. Additionally, the percentages shown above the bars represent the proportion of eventual successes that were achieved by a given attempt number. By definition, these values sum to 100% at the maximum number of attempts, as all queries that eventually succeeded are included.

The results show that for all models, the majority of achievable accuracy is obtained within the first attempt, with typically more than 85% of the successful queries achieved in the first attempt. Additional retries yield diminishing returns, while cumulative accuracy continues to increase, the proportion of new successes gained at later attempts is small, and very few successful queries require more than three attempts to converge. Differences across models are also apparent. Interestingly, the strongest model (*o3-mini-high)* reached 100% accuracy only at the 4$^{th}$ attempt (together with *microsoft/phi-4*, *qwen/qwen2.5-vl-72b-instruct*, gpt-*4.1-mini,* and *google/gemini-2.0-flash-001*). None of the models and queries solved at the 5$^{th}$ attempt, suggesting that increasing the allowed attempts is not expected to improve the performance. From a practical perspective, these findings are significant. For non-technical users, who may not be able to diagnose or correct Python error messages, the ability of the framework to automatically recover from execution errors is crucial. At the same time, the high proportion of first-attempt successes indicates that users will generally receive valid results quickly, without waiting through multiple debugging rounds. This balance between reliability and efficiency is especially important for potential deployment in utility settings, where timely decision support is essential.

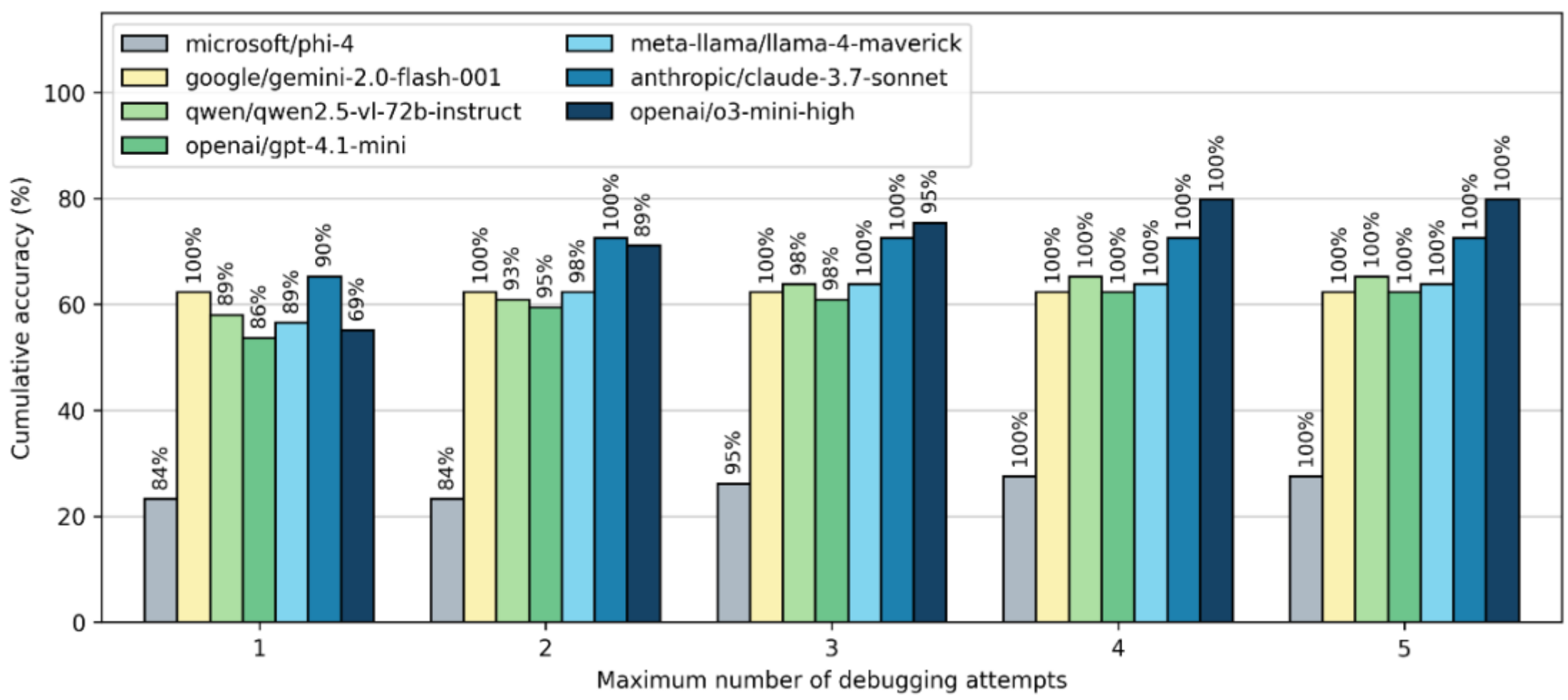

**Figure 6**: *Cumulative accuracy as a function of max debugging attempts by LLM. Cumulative accuracy as a function of the max number of debugging attempts. Bar heights indicate the cumulative accuracy. Percentages above the bars denote the proportion of successful queries out of total successful queries.*

## 3.2. Errors and Hallucinations Analysis

To better understand the limitations of the LLM-EPANET framework, the unsuccessful queries were analyzed based on two distinct categories: execution failures (where the generated code could not be executed) and incorrect results (where the code ran but produced an incorrect result). The main failure reasons (in both categories) were identified as follows: (1) Python Syntax - invalid Python structure or indentation that prevented execution. (2) Bad Logic - the generated code implemented an incorrect procedure. (3) Indexing - errors stemming from the discrepancies between EPyT indexing and EPANET conventions, such as off-by-one mistakes or out-of-bounds indices. While EPyT returns numpy arrays with indexing starting from 0, the EPANET index starts from 1. (4) Importing - incorrect import statements, often due to capitalization errors (e.g., import EPyT) or using calling packages without importing them (mostly numpy). (5) Aggregation - failures to correctly apply aggregation operators (mean, max, min) on EPyT outputs. (6) Signature Mismatch - the evaluation script is not calling the generated code correctly due to an error in the function name, number of arguments, or argument order. (7) EPyT Wrong Usage - misuse of EPyT functions or attributes not present in the package. (8) Other - a small number of miscellaneous cases.

A summary of the unsuccessful queries and their causes grouped by models is presented in Figure 7. Execution failures were dominated by Signature Mismatch and Importing issues. These failures typically occur when the code execution script generated by the small language model does not correctly import required libraries or calls the functions generated by the LLM with incorrect names or arguments. Although such errors do not reflect a misunderstanding of the underlying hydraulic problem, they pose a challenge for generating a robust interface between natural-language code generation and programmatic tool invocation. For most models, Python syntax errors were negligible. In contrast, wrong-answer failures correspond to cases where the generated code executes successfully but does not implement the correct modeling procedure or result extraction required by the query. The observed failure modes are distributed uniformly across three primary categories of Bad logic, Indexing issues, and Aggregation. These failure reasons indicate incomplete domain reasoning, such as applying an incorrect sequence of EPANET operations, struggling to extract the relevant information from simulation outputs, or aggregating results that do not answer the query.

Indexing which appeared as a frequent source of wrong answers, particularly due to the LLMs' confusion of EPANET's one-based indices with Python's zero-based arrays. Errors of this type led to systematically retrieving the wrong node or link values, producing plausible but incorrect outputs.

Aggregation mistakes were also observed, such as flattening arrays before applying argmax, or summing energy consumption across all links, although only the energy of pumps is asked. Figure 7 also reveals different error profiles across models. *Anthropic/claude-3.7-sonnet* and *o3-mini-high* exhibit more robust code generation, with the lowest number of execution failures. In contrast, *Microsoft/phi-4* fails to produce executable code in a substantial fractionof the cases (40 out of the 69 queries), indicating a limitation in reliable tool invocation. Across most models, Python syntax errors were observed with low frequency. Importing issues are prevalent in *google/gemini-2.0-flash-001* and gpt-*4.1-mini*. The results suggest that model capability primarily affects where failures occur, weaker models tend to fail at the execution level, while stronger models more often progress to runnable code but may still fail at the level of domain-specific reasoning.

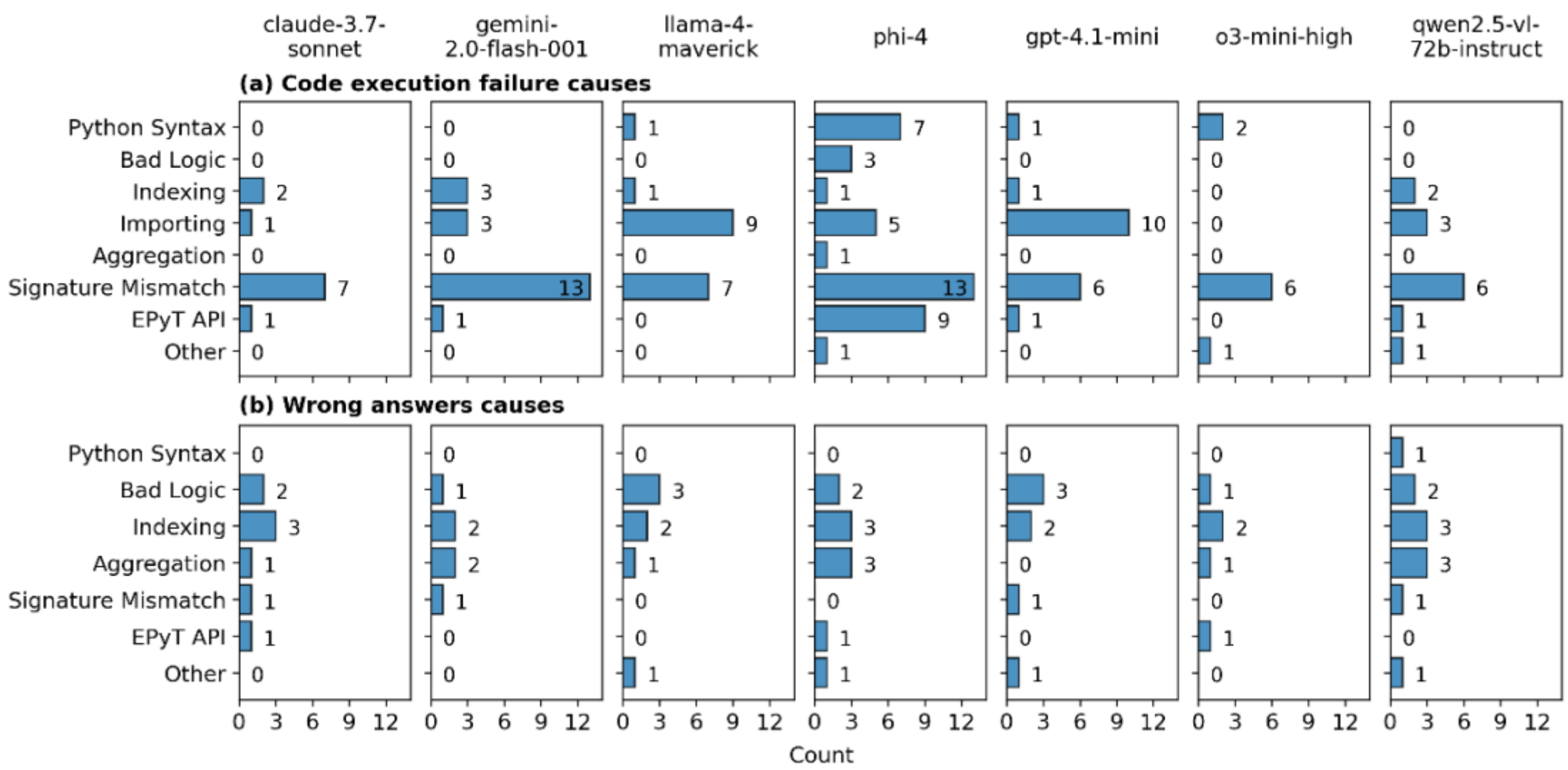

**Figure 7**: *Causes for unsuccessful queries. (a) Code execution failure causes, and (b) wrong answer failure causes, grouped by model.*

To complement the qualitative error taxonomy, Table 2 reports the frequency of execution-related and wrong-answer error categories, aggregated across all evaluated runs. The quantitative distribution is consistent with the patterns discussed in Figure 7, where execution errors are dominated by signature mismatches and importing-related issues, while wrong answer failures are primarily associated with indexing, aggregation, and logical reasoning errors. These error patterns provide guidance for future improvement of the framework. Frequent execution errors suggest that stricter function-signature validation and structured API schemas could reduce failures prior to execution. Wrong answer errors related to indexing, aggregation, and logic indicate the potential benefit of further prompt refinement or lightweight post-generation validation. More extensive approaches, such as model retraining or reinforcement learning, were not explored in this study and are left for future work.

**Table 2** - Frequency of error categories summary

| Category | Execution Errors (%) | Wrong Answers (%) |
|---|---|---|
| Signature Mismatch | 44.6 | 7.4 |
| Importing | 23.8 | 0 |
| EPyT API | 10 | 5.6 |
| Python Syntax | 8.5 | 1.9 |
| Indexing | 7.7 | 31.5 |
| Bad Logic | 2.3 | 25.9 |
| Other | 2.3 | 7.4 |
| Aggregation | 0.8 | 20.4 |

When discussing the framework errors, it is important to distinguish between code correctness and the hydraulic feasibility of the results. In the proposed LLM-EPANET framework, the LLM does not generate or approximate hydraulic physics, which are strictly enforced by the EPANET simulation engine. Consequently, when the LLM generates correct executable code, the resulting outputs are identical to those obtained by a human modeler using the same configuration. The framework does not currently attempt to judge whether a runnable solution is physically acceptable from an operational perspective, such as detecting negative pressures, tank bounds violations, or other infeasible operating conditions. These outcomes are meaningful diagnostic signals commonly encountered by human engineers during scenario analysis and design exploration. The LLM-EPANET does not try to explain hydraulic infeasibility. Rather, it exposes them transparently to the user. Such automated physical checks and validation rules can be considered for future work, particularly for decision-support applications involving non-expert users.

### 3.3. Prompt Analysis

As described in the methodology section, the framework relies on a meta prompt (see Section 2.6) that provides modelling guidance and syntax rules to the LLM. To assess the contribution of the meta prompt design to the LLM-EPANET performance, we conducted a prompt sensitivity analysis by comparing three levels of prompt complexity: Basic, Simple, and Complex designs of the meta prompts, which differ from each other in the level of detail of domain-specific modeling principles. All the prompts used in this study are attached to Appendix A.

Figure 8 presents the resulting accuracy across query categories for two representative models of *qwen/qwen2.5-vl-72b-instruct* and *openai/o3-mini-high*. The results show that increasing prompt structure leads to systematic performance improvements across all categories. The improvement is more pronounced in the complex categories, where the Basic and Simple prompts fail to answer correctly in almost all queries, while the complex prompt reaches above 60% accuracy. The observed effect of prompt complexity is highly consistent between the two models. In both cases, accuracy increases monotonically with prompt richness, and the relative gains are largest for categories requiring multi-step reasoning, such as Quality, Hydraulics Scenarios, and Iterative. For simpler categories, such as Static queries, the relatively small performance gap across prompt levels suggests that minimal instruction may already be sufficient. This is consistent with the expectation that such queries are closer to zero-shot tool usage, whereas the added value of the proposed framework becomes more evident for complex, multi-step tasks. For more complex queries, without a well-designed prompt, the framework's performance degrades substantially. Moreover, many of the failure types identified in our error analysis (e.g., aggregation mistakes, indexing errors) were mitigated by targeted instructions in the meta prompt, showing that careful prompt engineering is a practical strategy to enhance reliability in applied modelling tasks.

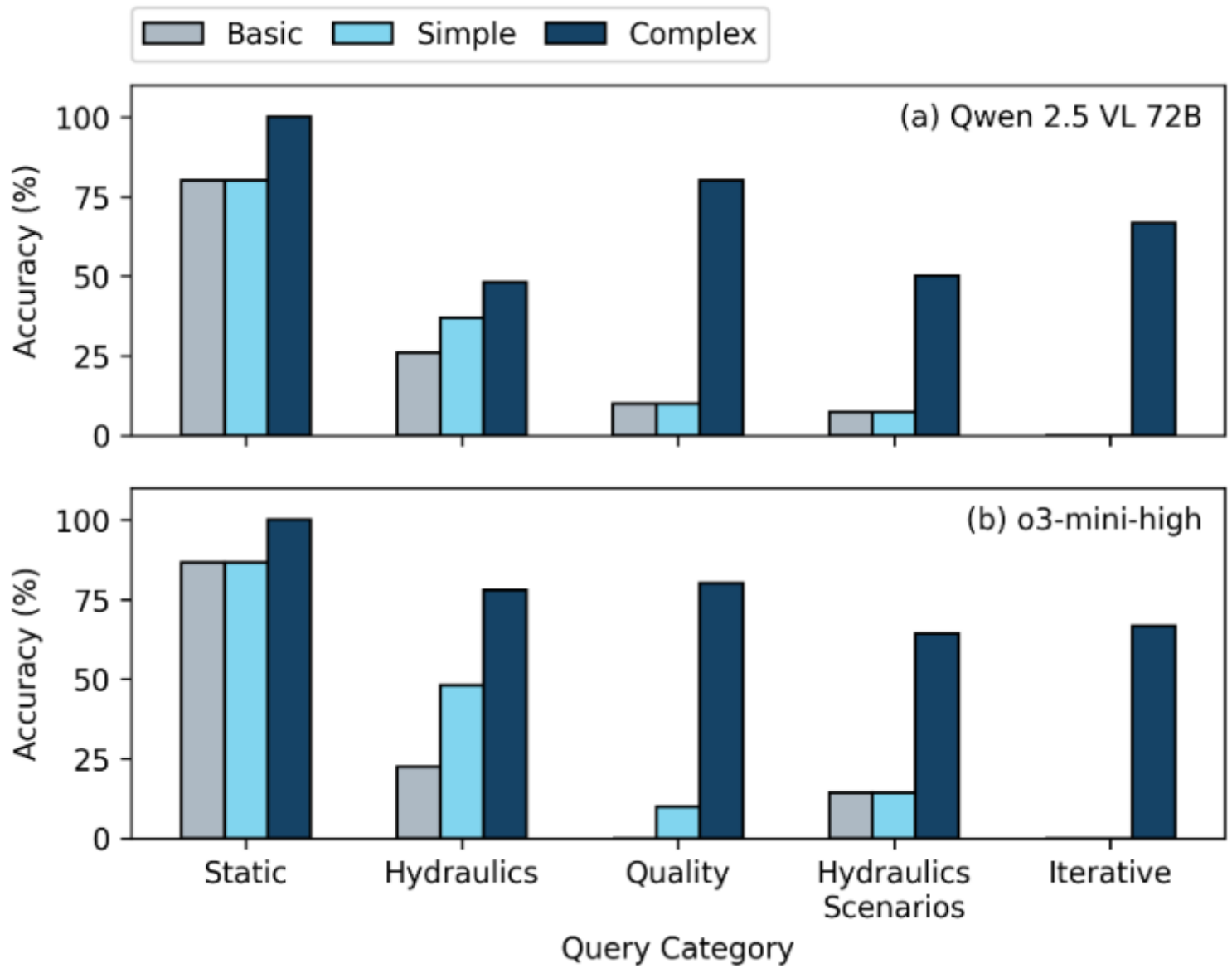

**Figure 8**: *Meta prompt impact on accuracy for openai/o3-mini-high and qwen/qwen2.5-vl-72b-instruct*

### 3.4. <u>Accuracy-Cost Tradeoff</u>

To further compare model performance, we analyzed both the cost and execution time latency of LLM usage, which are critical factors for embedding the proposed framework into practical workflows in the water sector. The selection of models for this study intentionally spans a wide range of pricing levels, from low-cost models such as *google/gemini-2.0-flash-001*, to models at the top of the pricing spectrum, such as *anthropic/claude-3.7-sonnet* (based on market pricing and model availability during the study period, late 2025 to early 2026). Figure 9 illustrates the tradeoffs between model accuracy, total cost, and execution time. The left plot (a) presents the accuracy-cost relationship, showing that higher accuracy is generally associated with increased cost, although the relationship is not monotonic. A Pareto-efficient region emerges in the high-accuracy, low-cost portion of the plot, where several models provide competitive accuracy at relatively modest cost. In particular, *google/gemini-2.0-flash-001, qwen/qwen2.5-vl-72b-instruct*, *meta-llama/llama-4-maverick*, and *openai/gpt-4.1-mini* achieve accuracy levels comparable to more expensive models while incurring substantially lower cost. The highest accuracy is achieved by *openai/o3-mini-high*, which correctly answers 80% of the queries. Nevertheless, this accuracy comes with a price, where the total cost is summed to $1, which is more than 4 times the cost of the above efficient models.

The right plot (b) presents the average response time per successful query as a function of the overall accuracy. The results suggest an accuracy-latency tradeoff, with higher accuracy models exhibiting longer response times. However, this relationship is not strictly linear or monotonic. Notably, the *qwen/qwen2.5-vl-72b-instruct* model achieves accuracy comparable to several mid-accuracy models (*meta-llama/llama-4-maverick*, *openai/gpt-4.1-mini*, and *google/gemini-2.0-flash-001*), yet its response times are slightly longer, also compared to that of *anthropic/claude-3.7-sonnet*, which achieves higher overall accuracy. The *openai/o3-mini-high* shows substantially longer response times than all other evaluated models, highlighting that its high accuracy comes at the cost of increased user waiting time. Taken together, these cost and latency results demonstrate that no single model dominates across accuracy, cost, and latency dimensions. Instead, model selection involves explicit trade-offs that depend on user and application requirements.

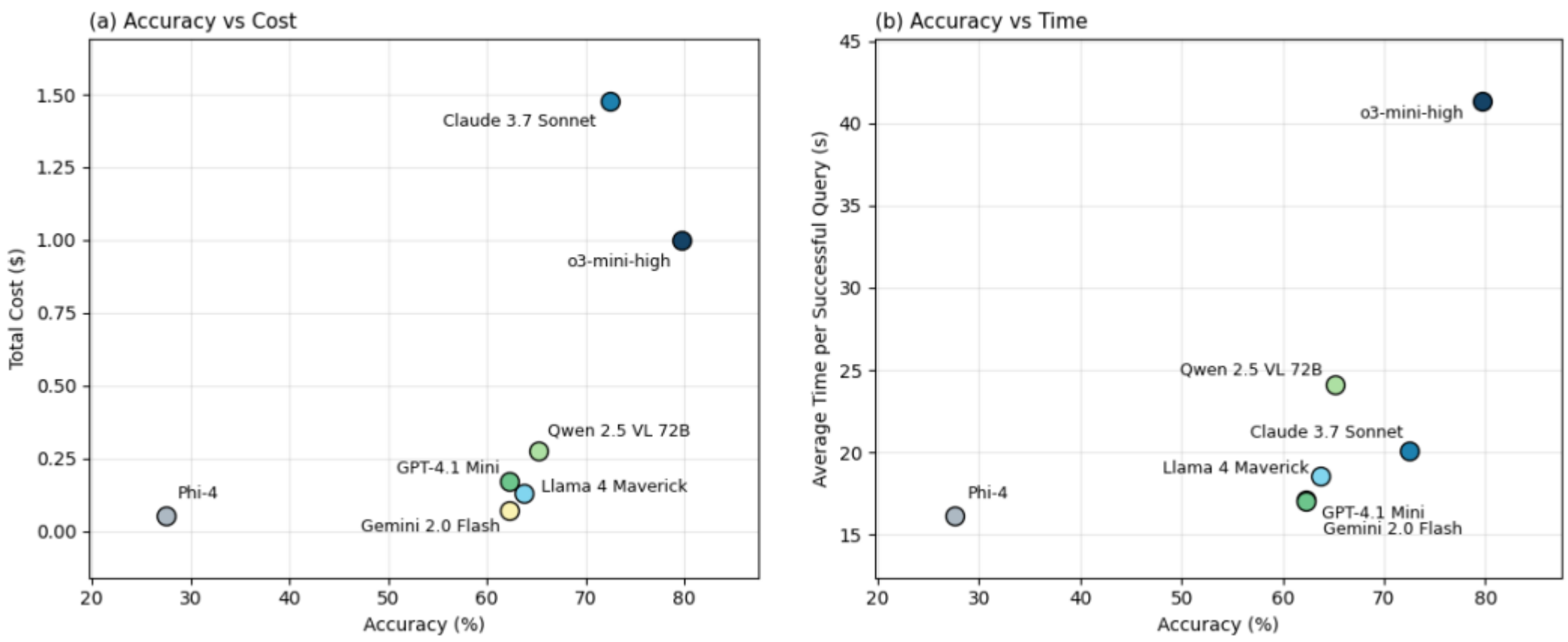

**Figure 9**: *LLMs cost and latency analysis*

## 4. Discussion

AI and LLM technologies are developing at an accelerated pace, reshaping technical workflows. The water sector, as a data-rich but traditionally conservative field, should be aware of these changes and consider adaptation. The integration of AI-driven interfaces into simulation tools such as the proposed LLM-EPANET marks a crucial step toward making advanced modeling more accessible and impactful across a broader range of users.

Given the rapid AI development, it is likely that by the time this paper is published, the underlying technologies will have already evolved significantly. New paradigms such as Model Context Protocols (MCP) are emerging, promising further streamlining of human-software interaction through structured, context-aware communication. Similarly, the increasing maturity of AI agents is expected to enable not only natural language interaction but also the autonomous execution of multi-step complex tasks such as network optimization, fault detection, and scenario response planning. These advancements point toward a future where simulation, analysis, and decision support are integrated into cohesive, intelligent workflows.

However, while embracing innovation is critical, it must be accompanied by a measured and responsible approach to adoption. LLMs and generative AI models are inherently stochastic; they can produce impressive results, but are also prone to hallucinations and errors, especially in domain-specific contexts where training data is limited. Thus, the integration of AI into water systems engineering should aim to augment, not replace, expert knowledge. Experts play a vital role not only in validating model outputs but also in defining the assumptions, objectives, and constraints that govern responsible system design and operation. Expanding access to simulation tools through natural language interfaces offers an opportunity to amplify expert capabilities, rather than replacing them. By lowering the entry barrier, these interfaces allow non-technical stakeholders to engage more actively in modeling and analysis, contributing to more transparent workflows and increased model-based decision validation. This democratization of technical tools can elevate the overall quality of discussions, enhance operational efficiency, and ultimately lead to more resilient, reliable, and cost-effective water services.

In conclusion, the water sector should actively engage with the accelerating landscape of AI technologies. By leading their responsible adoption, professionals can utilize these tools to serve the core mission of delivering reliable and high-quality water services to the public.

## 5. Conclusions

This paper presents the development of an LLM-based framework for natural language interactions with the WDS simulator software EPANET. The proposed agent-driven approach converts natural language queries into executable code that calls EPANET toolkit and returns the code output as results. The LLM-EPANET framework is a systematic approach that combines LLM code generation capabilities with the modeling tasks faced by professionals in the water sector. The aim is to democratize the use of sophisticated computational tools, extend their user base, and enhance data-driven management of WDS.

The proposed framework was evaluated using a set of 69 queries spanning five categories of complexity and across several benchmark water networks and commercially available LLMs. The results demonstrate that LLMs can effectively serve as intermediate modelers and support modeling tasks that are often inaccessible to non-technical users and consume time and effort from expertise. The best performing LLMs achieved overall accuracies between 56% and 81%, with above 90% accuracy for simpler query types.

Looking ahead, the rapid evolution of AI continues to unlock new opportunities for human-software interactions, opening vast research and commercial software development opportunities in the water sector. Future research should investigate not only improved prompt engineering and automation strategies but also the interpretability and explainability of simulation and optimization results. By embracing these developments, the water sector can harness AI technologies to enhance operational efficiency, improve decision transparency, and ultimately better serve public needs.

## 6. Declaration of generative AI and AI-assisted technologies in the manuscript preparation process

During the preparation of this work, the authors used ChatGPT to refine argumentation and improve the writing style in part of the manuscript. The authors reviewed and edited the generated content as needed and take full responsibility for the content of the published article.

## 7. Appendix A - Prompts

```
basic_prompt = """
You are a code assistant for a water engineer.
Your task is to write code snippets that interact with the EPyT
python package based on its documentation,
and perform the task needed.

Task: Write a python function that fulfills the request of < {task} >

These are relevant functions from the EPyT library retrieved from the
documentation:

<CONTEXT_START>

{context}

<CONTEXT_END>

A python function that returns the request of < {task} > is:
"""
```

```
simple_prompt = """
[INSTRUCTIONS_START]

You are a code assistant for a water engineer.
Your task is to write code snippets that interact with the EPyT
python package based on its documentation,
and perform the task needed.
Unless written otherwise, the code snippets should return the needed
output in a return statement.
For any calculations needed, you need to run a simulation and get the
results from the simulation.
Write the code needed for the following task only, without any
additional text or comments.
If the given task is a query for a specific parameters value, you
need to add the parameters as arguments to the function.
If the given task is a query for a specific function, you need to
write a function that returns the function.

[INSTURCTIONS_END]

Task: Write a python function that fulfills the request of < {task} >

These are relevant functions from the EPyT library retrieved from the
documentation:

[CONTEXT_START]

{context}

[CONTEXT_END]

A python function that returns the request of < {task} > is:
"""
```

```
complex_prompt = """
[INSTRUCTIONS_START]

# Ovrerview

You are a code assistant for a water engineer.
Your task is to write code snippets that interact with the EPyT
python package based on its documentation,
and perform the task needed.

Unless written otherwise, the code snippets should return the needed
output in a return statement.

For any calculations needed, you need to run a simulation and get the
results from the simulation.

Write the code needed for the following task only, without any
additional text or comments.

If the given task is a query for specific parameter values, you need
to add the parameters as arguments to the function.

If the given task is a query for a specific function, you need to
write a function that returns the function.

# Some details you should know about EPyT:

1) You can use builtin functions of EPyT to run complete hydraulics
and quality simulations, for example:
    d.getComputedTimeSeries()
    d.getComputedQualityTimeSeries()
    These options generates all the required results, and in some
cases easier than writing your own loops for simulation
    avoid using the function: d.runHydraulicAnalysis(), it is used to
run only single period analysis and not full simulation

    In queries regarding hydraulic parameters such as pressure, flow,
headloss use the following:
    d.getComputedTimeSeries().Pressure,
d.getComputedTimeSeries().Flow, d.getComputedTimeSeries().HeadLoss

    In queries regarding water quality, the results will always be
stored under NodeQuality
    For example if you are asked about water age, chlorine
concentration, or portion of water source use:
    d.getComputedQualityTimeSeries().NodeQuality

2) Indexing
    EPANET uses 1-based indexing; Python uses 0-based indexing.
    Adjust indices as needed:
    Python to EPANET: `epanet_index = python_index + 1`
    EPANET to Python: `python_index = epanet_index - 1`

3) Time Step Reporting
    Results are generated at uniform intervals defined by the report
time step parameter.
    When retrieving results at time t (in hours since the start of
the simulation),
    obtain the report time step in seconds (dt =
```

```
d.getTimeReportingStep())
    and compute the reporting index as t_idx = int(round((t * 3600) /
dt))

4) Energy Calculations
    Energy or power refers to the energy consumed by pumps only.
    You will need to extract the indexes of pumps for the energy
consumption computation such as:
    d.getComputedHydraulicTimeSeries().Energy[:, [_ - 1 for _ in
d.getLinkPumpIndex()]]

5) Water Quality Simulations
    These start with zero initial water quality.
    Ensure simulation duration is at least 168 hours.

6) Element IDs and Indices
    In EPyT (and EPANET) every element has an index (int) and ID
(str).
    The queries by users will be based on element IDs. Nothe that ID
can be numeric, but it is still ID and not index.
    Convert these to indices using EPyT functions. Examples to
conversions functions:
    To get the index of a link (pipe, pump, valve) use:
d.getLinkIndex(link_id)
    To get the index of a node (junction, reservoir, tank) use:
d.getNodeIndex(node_id)
    Use other functions if necessary

    When using EPyT functions, note whether the input should be Index
or ID

    The user expect to get answers base on the ID of elements and not
index
    Use EPyT to convert from index to ID when needed, for example:
    d.getNodeNameID('<element_int_id>')

    The EPANET indexing is sequential increasing integers starting
from 1
    Nodes and Links are indexed separately.
    Node is a class of all the elements that can be represented as
node - junctions, reservoirs, tanks
    Link is a class of all the elements that can be represented as
link - pipes, pumps, valves
    Nodes:  Junctions → Reservoirs → Tanks
    Links:  Pipes → Pumps → Valves

7) Note that tank levels are represented as the pressure in the
tanks. Use EPyT Pressure property to answer queries
regarding tank level

8) If you need to ignore reservoirs and tanks you can eliminate the
last d.getNodeReservoirCount() + d.getNodeTankCount() elements
For example to get pressure in junctions while ignoring reservoirs
and tanks use:
d.getComputedTimeSeries().Pressure[:, :-(d.getNodeReservoirCount() +
d.getNodeTankCount())]

9) If you are asked about demands in the network, ignore the demands
of reservoirs and tanks. You can use:
d.getComputedTimeSeries().Demand[:, :-
(d.getNodeReservoirCount()+d.getNodeTankCount())]
```

````
10) Avoid from trying to close the epanet instance we method like
d.close(). The epanet will be closed automatically
once the program is finished

11) If you need to close a pipe, you need to set its initial status
with this command:
d.setLinkInitialStatus(link_index, 0) # 0 for closed, 1 for open

# Assumptions and Clarifications

1) Assume an import of `epanet from epyt` is present and  code line
that initiates the epanet model object is always present before the
function definition: `d = epanet('Net1.inp')`

2) The function always takes `d` as the first argument:
```python
def func_name(d, arg1, arg2, ...):
    ...
```

[INSTURCTIONS_END]

Task: Write a python function that fulfills the request of < {task}
>.
Write the code needed for the following task only, without any
additional text or comments.


[CONTEXT_START]

These are relevant functions from the EPyT library retrieved from the
documentation:

{context}

[CONTEXT_END]

A python function that returns the request of < {task} > is:
"""
````

```
code_minor_fix_prompt = """
[INSTRUCTIONS_START]

You are a code assistant.
You are given a code snippet that is a function definition that uses
a water engineering python framework called EPyT.
If needed, do the following changes to the code snippet:

1) If the function definition does not have a return statement, add a
return statement that returns None.
2) If the function definition does not use have an epanet model
object `d` as the first argument, add it as the first argument and
delete any other `d` variables defined in the function.
3) If the function definition uses `d.getSomeFunctionIndex(some_id)`,
then replace it with `d.getSomeFunctionIndex(str(some_id))`.

Do not add any notes or comments to the code you are generating.
```

```
[INSTRUCTIONS_END]

The fixed code snippet is:
"""
```

```
code_exec_prompt = """
<instructions>
You are a model that takes a code snippet as input and generates an
executable code one-line that calls upon the code snippet to return
the result of the given task.
The task was given to you by a water engineer, and uses the EPyT
python package.
The EPyT package is imported as epyt.
Unless written otherwise, the first argument `d` is the EPyT epanet
model object.
Function always takes `d` as the first argument.
Do not add any notes or comments to the code you are generating.
</instructions>

Your task is given:

{task}

The code snippet is:

[CODE_SNIPPET_START]

{context}

[CODE_SNIPPET_END]

<instructions>
You should write a one line code that fulfills the request of the
given task, with the given arguments.
Your output should be a one-liner that completes the given task using
the given code snippet.
The one liner should be in the form of:
func_to_call(d, arg1, arg2, ...)
where `func_to_call` is the function that is called in the code
snippet, and `arg1, arg2, ...` are the arguments needed for the
function.
Unless written otherwise, the always-needed function argument `d` is
the EPyT epanet model object. This argument should always be the
first argument in the snippet.
You are given the arguments needed for the function call.
Do not add any notes or comments to the code you are generating.
Do not add any quotes around the one-liner of any kind.
</instructions>

The arguments are: < {args} >

An example of a one-liner that calls the code snippet is:
"""
```

```
code_exception_handling = """
<instructions>
You are given a code snippet that is a function definition that uses
a water engineering python framework called EPyT, which is imported
```

````
as epyt.
This code snippet raises an exception when called, and you need to
fix the exception.
The code snippet consists of an import statement, a epanet model
object creation, a function definition, a one liner that calls the
function, and a print statement.
The function definition or the one liner is the one that raises the
exception.
If the exception is "Exception: 'int' object has no attribute
'encode'", then see if `d.getNodeIndex(node_id)` is used, and replace
it with `d.getNodeIndex(str(node_id))`.
In your response, provide the fixed code snippet only without any
additional text.

Example of the code snippet structure that you should return:
```
from epyt import epanet
d = epanet('Net1.inp')

def func_name(d, arg1, arg2, ...): # use the args needed for the
function, d is always present and the first argument
    func_code
    return return_val
return_val = eval_script
print(return_val)
```
</instructions>

The code snippet is:

[CODE_SNIPPET_START]

{code}

[CODE_SNIPPET_END]

The exception raised is:

[EXCEPTION_START]

{exception}

[EXCEPTION_END]


1. If the function definition does not have a return statement, add a
return statement that returns None.
2. If the function definition does not use have an epanet model
object `d` as the first argument, add it as the first argument and
delete any other `d` variables defined in the function.
3. If the function definition uses `d.getSomeFunctionIndex(some_id)`,
then replace it with `d.getSomeFunctionIndex(str(some_id))`.
4. Make sure that you have to change the code snippet in some way to
fix the exception.

A fixed code snippet that does not raise the exception is:

"""
````

```
code_review_prompt = """
You are a code reviewer for water engineering code that uses the EPyT
framework.
You are given:
1. The original task, which is a query from a water engineer
2. The generated function code, meaning a function with a def
statement and a return statement
3. The generated evaluation code, meaning a one-liner that calls the
function with the needed arguments

The original task is:
< {task} >

Your task is to review the code and check if it follows these
requirements:
1. The function must have 'd' as its first parameter
2. All node/link IDs passed to EPyT functions must be strings (e.g.,
d.getNodeIndex(str(node_id)))
3. The function must have a return statement
4. The evaluation code must be a single line calling the function
5. Time values in the output should be in hours (divide by 3600 if
needed)
6. Results should use element IDs and not indices in the output
7. The function name should be descriptive of its purpose
8. There should be no imports in the function code
9. There should be no network loading in the function code (e.g d =
epanet(inp_path))

Respond with a JSON object that adheres to this schema:
{review_result_schema}

If a requirement is not met, set "needs_revision" to true and specify
the issues in the respective lists.
If all requirements are met, set "needs_revision" to false and both
issue lists to empty.
If a requirement is not relevant to the provided code, you can ignore
it.

Generated Function:
[FUNCTION_START]
{function_code}
[FUNCTION_END]

Evaluation Code:
[EVAL_START]
{eval_code}
[EVAL_END]

Review result:
"""
```

**Large Language Models for Water Distribution Systems Modeling and Decision-Making – Supplementary Material**

1. **Retrieval Ablation.**

To assess the sensitivity of the retrieval component to alternative search strategies, we conducted an ablation study comparing the semantic retrieval baseline with hybrid lexical–semantic retrieval variants under different mixing parameters (α), where *α=0* denotes pure lexical search. The analysis evaluates both average retrieval quality and robustness across queries as a function of the number of retrieved documents (*k*). This supplementary analysis is provided for completeness and does not alter the main conclusions or design choices adopted in the paper.

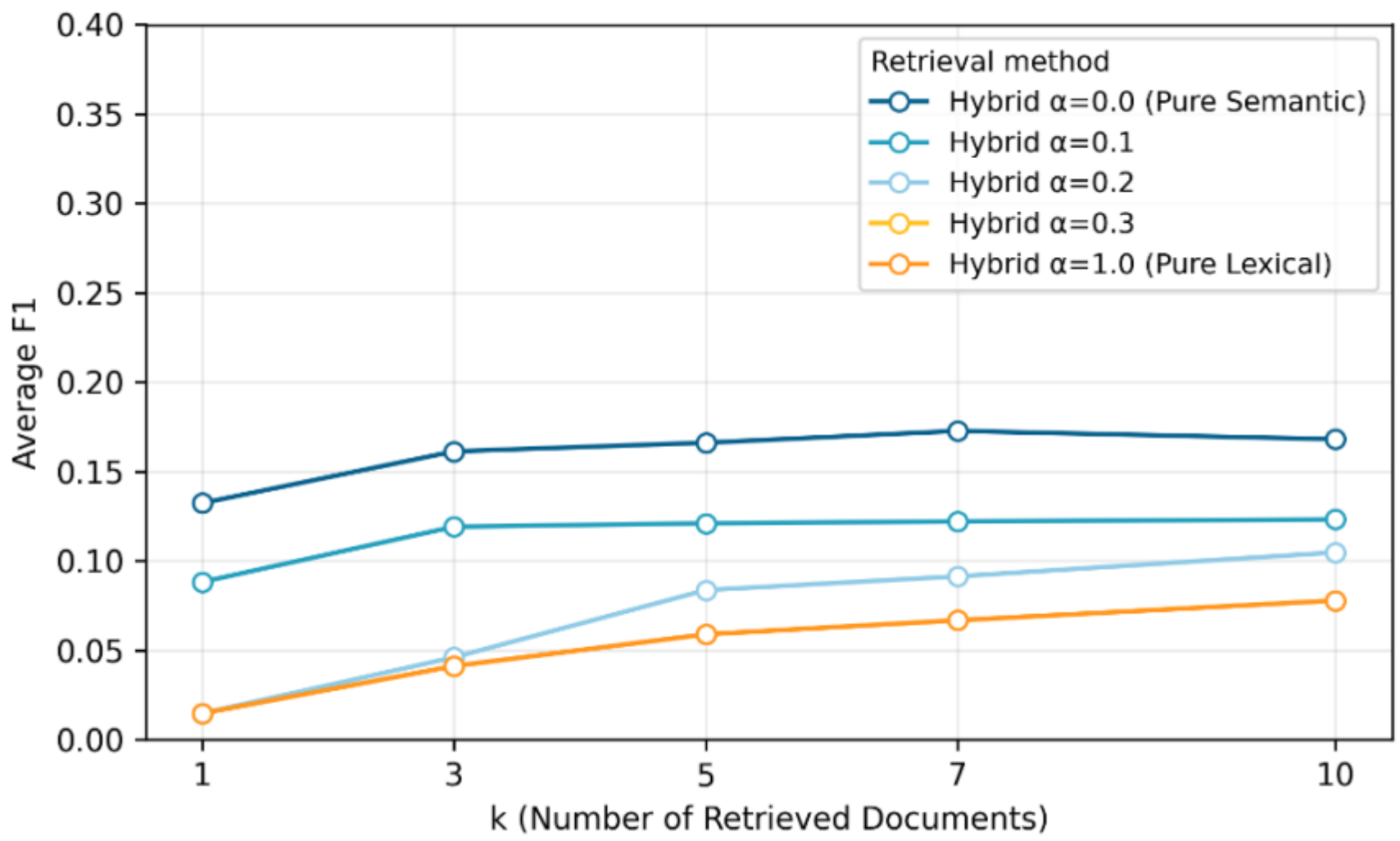

**Figure S1.** *Mean retrieval performance across search strategies.*

Figure S1 presents the mean F1 score as a function of the number of retrieved documents (*k*) for the pure semantic retrieval (baseline) and hybrid lexical–semantic retrievals with different mixing parameters (*α*). Results are averaged across all evaluation queries. While hybrid retrieval variants exhibit different trends depending on *α*, no configuration consistently outperforms semantic retrieval across all values of *k*, clarifying why semantic retrieval was adopted as the default configuration in the main analysis.

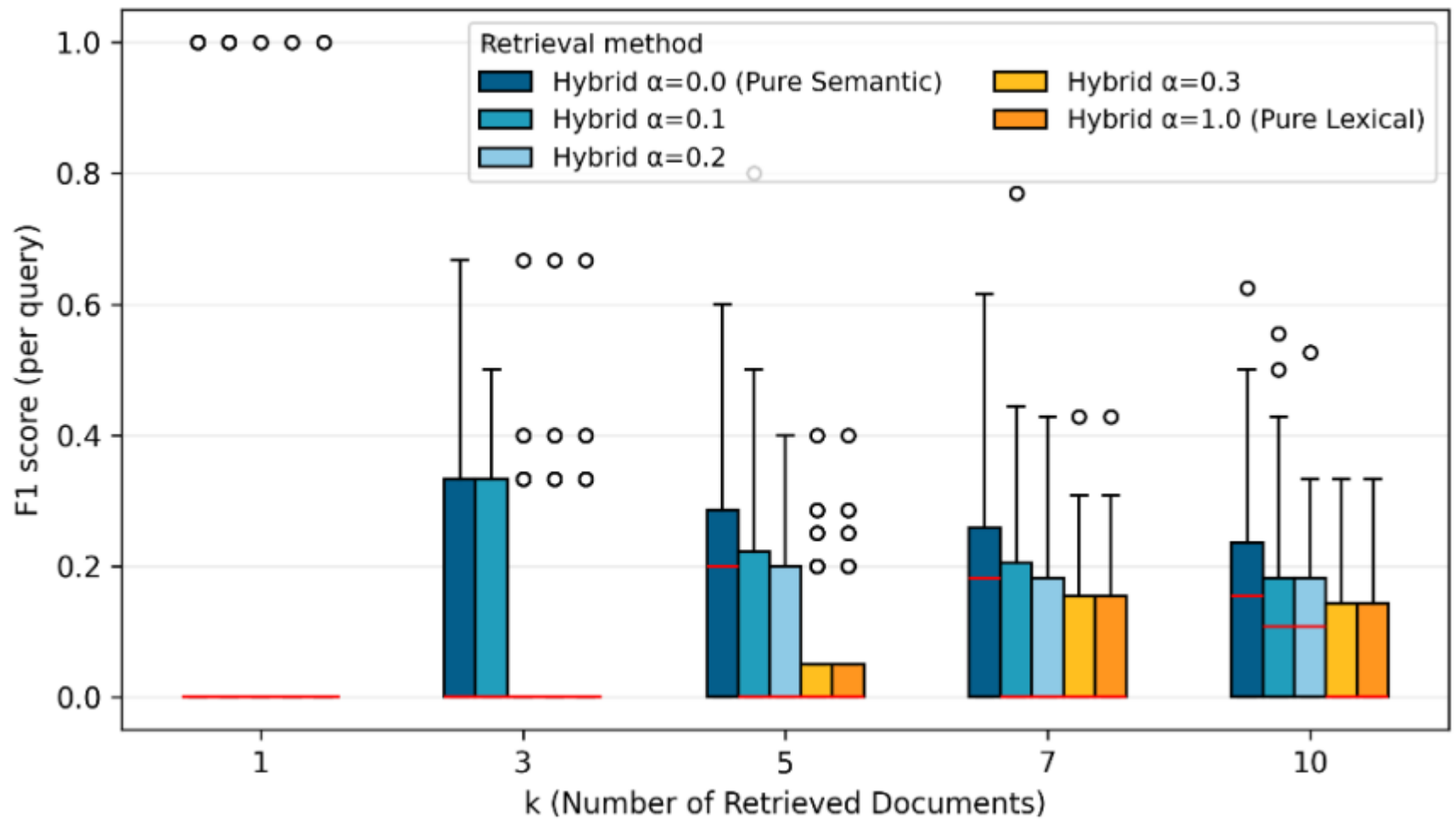

**Figure S2.** *Distribution of per-query retrieval performance across search strategies.*

Figure S2 presents the distribution of per-query F1 scores for semantic and hybrid retrieval methods at different values of *k*. The results highlight substantial variability across queries, particularly for small *k* values, and show that hybrid retrieval does not systematically improve robustness relative to semantic retrieval, and in most cases, the hybrid search is inferior to semantic.

## 2. Sensitivity Analysis of Number of Retrieved Documents.

To assess the sensitivity of the framework to the number of retrieved documents, we evaluated end-to-end task accuracy as a function of Top-K for two representative large language models. This analysis is provided to justify the choice of retrieval parameters and does not affect the main conclusions of the paper.

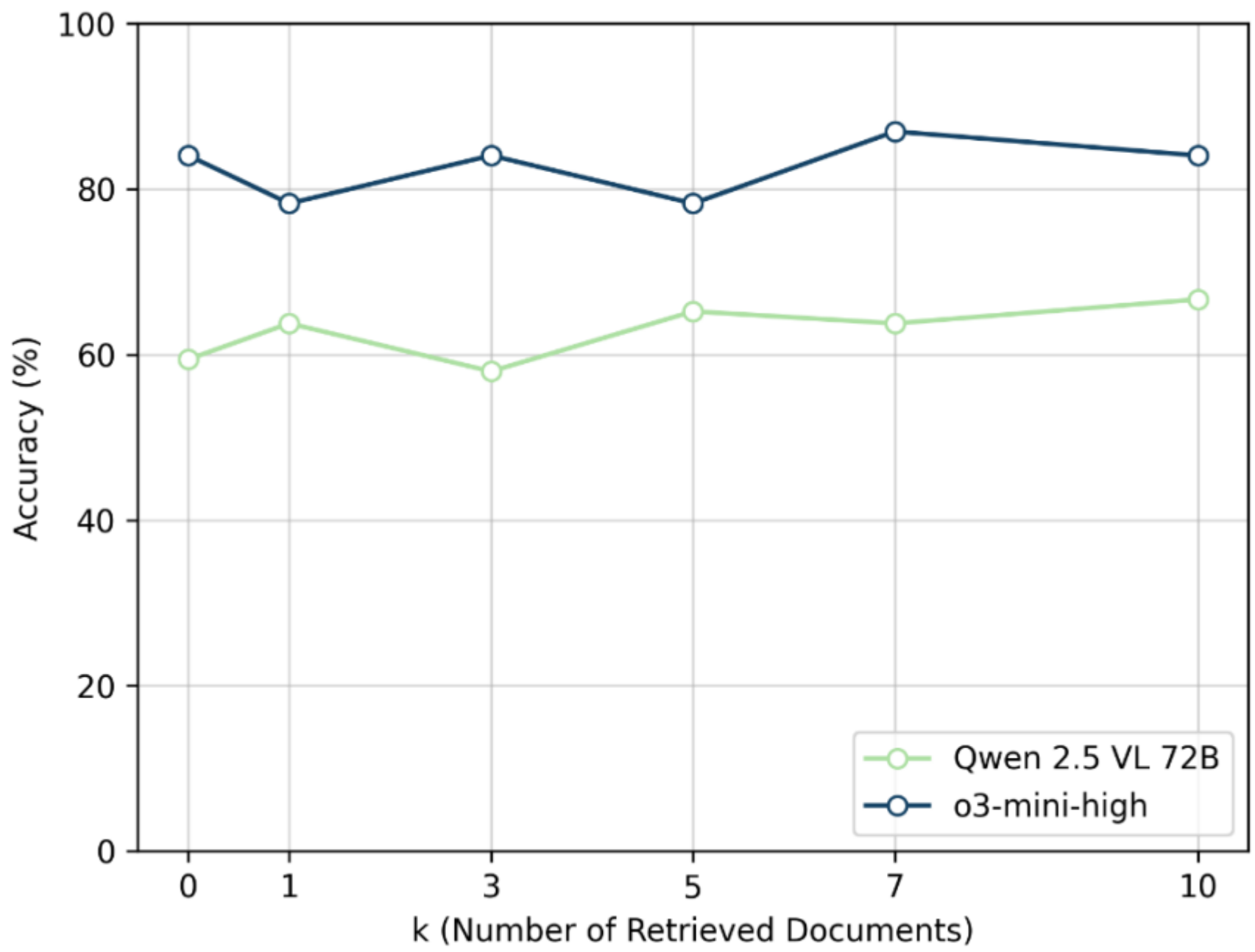

**Figure S3.** *Sensitivity of end-to-end performance to the number of retrieved documents (Top-K).*